\documentclass[10pt,twocolumn]{article}
\usepackage{graphicx}
\usepackage[table]{xcolor}
\usepackage{amssymb,amsthm,amsmath}
\usepackage[
  letterpaper,
  columnsep=0.375in,
  textwidth=6.875in,
  textheight=9.25in,
  top=0.75in,
  bottom=0.75in,
  left=0.75in,
  right=0.75in,
  headheight=12pt,
  headsep=0.25in,
  footskip=0.3in
]{geometry}
\usepackage{breakurl}

\usepackage{lineno}               
\usepackage{float}
\usepackage{multirow}
\usepackage{booktabs}
\usepackage{tabularx}
\usepackage{csquotes}
\usepackage{paralist}
\usepackage{hyperref}
\usepackage{fancyhdr}
\usepackage{etoolbox}
\usepackage{authblk}
\usepackage{caption}
\newcommand{\eref}[1]{Eq.~(\ref{#1})}
\newcommand{\fref}[1]{Fig.~\ref{#1}}
\newcommand{\ie}{\emph{i.e.}}
\newcommand{\eg}{\emph{e.g.}}
\newcommand{\expv}[1]{\ensuremath{\langle #1 \rangle}}
\makeatletter
\def\section{\@startsection{section}{1}{\z@}{-3.25ex plus -1ex minus -.2ex}{1.5ex plus .2ex}{\normalfont\large\bfseries\centering}}
\makeatother
\hypersetup{
  colorlinks=true,
  linkcolor=blue,      
  filecolor=black,
  urlcolor=blue,
  citecolor=blue       
}

\title{Universal computational thermal imaging overcoming the ghosting effect} 
\author[1]{Hongyi Xu}
\author[2]{Du Wang}
\author[3]{Chenjun Zhao}
\author[1]{Jiashuo Chen}
\author[6]{Jiale Lin}
\author[4*]{Liqin Cao}
\author[2]{Yanfei Zhong}
\author[5*]{Yiyuan She}
\author[1,6,7,*]{Fanglin Bao}

\affil[1]{Department of Electronic and Information Engineering, School of Engineering and Research Center for Industries of the Future, Westlake University, Hangzhou 310030, China}
\affil[2]{State Key Laboratory of Information Engineering in Surveying, Mapping, and Remote Sensing, Wuhan University, Wuhan 430079, China}
\affil[3]{Changchun Institute of Optics, Fine Mechanics and Physics, Chinese Academy of Sciences, Changchun 130033, China}
\affil[4]{School of Resource and Environmental Sciences, Wuhan University, Wuhan 430079, China}
\affil[5]{Institute of Theoretical Science, Westlake University, Hangzhou 310030, China}
\affil[6]{Department of Physics, School of Science, Westlake University, Hangzhou 310030, China}
\affil[7]{Institute of Natural Sciences, Westlake Institute for Advanced Study, Hangzhou 310024, China}

\affil[*]{clq@whu.edu.cn, sheyiyuan@westlake.edu.cn, baofanglin@westlake.edu.cn}

\date{\today}

\begin{document}
\maketitle
\begin{abstract}
Thermal imaging is crucial for night vision but fundamentally hampered by the `ghosting effect' \cite{lawson1956implications, schmidt2014thermal, Rafael2021}, a loss of detailed texture in cluttered photon streams \cite{Bao2024}. While conventional ghosting mitigation has relied on data post-processing, the recent breakthrough in heat-assisted detection and ranging (HADAR) \cite{Bao2023} opens a promising frontier for hyperspectral computational thermal imaging that produces night vision with day-like visibility \cite{Pile2023, TEFormer2025, XU2025106114, Xinyu2025}. However, universal anti-ghosting imaging remains elusive, as state-of-the-art HADAR applies only to limited scenes with uniform materials, whereas material non-uniformity is ubiquitous in the real world \cite{Jianrui2021}. Here, we propose a universal computational thermal imaging framework, TAG (thermal anti-ghosting), to address material non-uniformity and overcome ghosting for high-fidelity night vision. TAG takes hyperspectral photon streams for nonparametric texture recovery, enabling our experimental demonstration of unprecedented expression recovery in thus-far-elusive ghostly human faces — the archetypal, long-recognized ghosting phenomenon. Strikingly, TAG not only universally outperforms HADAR across various scenes, but also reveals the influence of material non-uniformity, shedding light on HADAR's effectiveness boundary. We extensively test facial texture and expression recovery across day and night, and demonstrate, for the first time, thermal 3D topological alignment and mood detection. This work establishes a universal foundation for high-fidelity computational night vision, with potential applications in autonomous navigation, reconnaissance, healthcare, and wildlife monitoring.

\end{abstract}

\begin{figure*} 
    \centering
    \scalebox{1}{\includegraphics{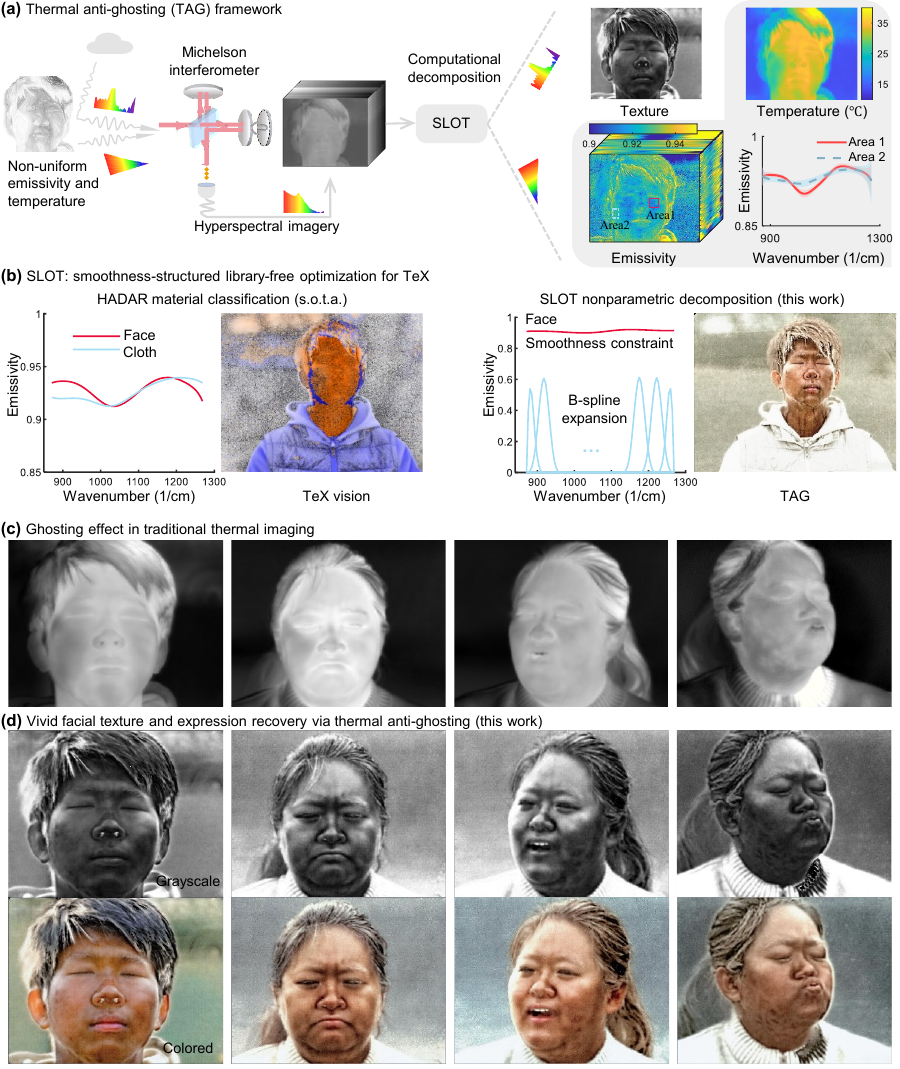}}
    \caption{Universal thermal anti-ghosting (TAG) for high-fidelity night vision with day-like visibility. a, The TAG framework. TAG takes hyperspectral thermal imagery for nonparametric TeX decomposition and produces high-fidelity textures, along with non-uniform temperature and emissivity. b, The SLOT principle. To break TeX degeneracy, SLOT uses a B-spline basis expansion for continuous emissivity and imposes a smoothness constraint, universally applicable to non-uniform materials. In contrast, HADAR relies on rigid material categorization and suffers from classification errors when material non-uniformity is present. c, Traditional thermal imaging cannot separate ambient scattering from direct emission, yielding textureless silhouettes that resemble ghosts. d, TAG vividly recovers otherwise hidden geometric textures and facial expressions that are crucial for various perception tasks. From left to right: neutral (sad), frown with eyes open (angry), grin with eyes and mouth open (happy), and pout.}
    \label{fig:fig1}
\end{figure*}

\section{Introduction}
\label{sec:introduction}
Night vision is a long-sought fantastical capability that enables perception in pitch darkness. Since the discovery of infrared light in 1800 \cite{Herschel1800}, advances in infrared sensing have established thermal imaging as a practical route to night vision and a vital tool for machine perception \cite{Gade2014, ROGALSKI2011136, Hwang2015, Sun2019}. Such imaging allows, for instance, pedestrian detection in autonomous navigation \cite{Alejandro2016, GUAN2019, Zhou2020, Chio2018, Khattak2020}, yet it typically yields textureless, ghostlike silhouettes (see \fref{fig:fig1}c for example) that obstruct critical perception tasks, including facial recognition \cite{Treible_2017_CVPR, Riggan2018}, stereo reconstruction \cite{Li_2023}, and semantic segmentation \cite{Ha2017, Shivakumar2020}. To mitigate this ghosting effect, traditional approaches have turned to image post-processing algorithms for visual contrast enhancement (see, \eg, CLAHE and its variants \cite{Yash2021, Soundrapandiyan2022, Dhal_2021, LI_2018}; and machine learning techniques \cite{Bouhlel_2023, Pang_2023, KUANG_2019, Lee_2017}). Although important strides have been made \cite{Hou2025, Chen2025, Wen2025}, ghosting — and the attendant loss of irrecoverable visual information — remains a major hurdle to thermal perception \cite{Rafael2021, beaudry2012intuitive}, restricting the seamless operation of artificial intelligence (AI) agents across day and night.
\begin{figure*}[t!]
    \centering
    \scalebox{1}{\includegraphics{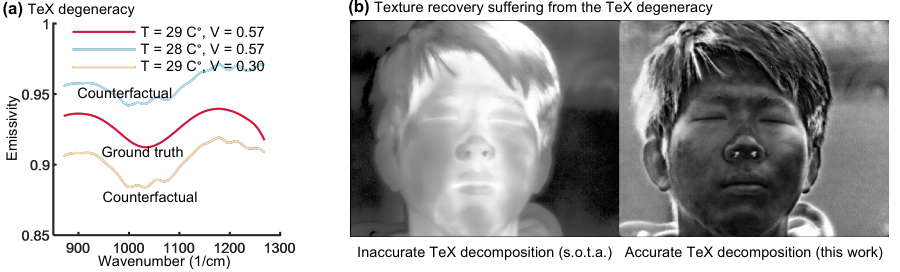}}
    \caption{The challenge of TeX degeneracy for texture recovery. a, Illustration of TeX degeneracy, where multiple TeX triplets produce an identical thermal radiation spectrum (taken from the red cross in b). b, Texture comparison. Left: poor texture recovery by panchromatic thermal imaging and CLAHE failing to break the TeX degeneracy. Right: vivid texture recovery by TAG, which successfully breaks the TeX degeneracy.}
    \label{fig:fig2}
\end{figure*}

Only recently, the ghosting mechanism was revealed \cite{Bao2024} as the degeneracy among the physical attributes of temperature $T$, emissivity $e$, and texture $X$ in thermal radiation. When thermal photons are emitted and scattered, the geometric texture $X$ of the scene is irreversibly lost in panchromatic thermal images, since there is an infinite number of TeX solutions that can lead to the same observed signal. Remarkably, by leveraging hyperspectral thermal imaging and unveiling the key of a pre-calibrated `material library', HADAR \cite{Bao2023} breaks TeX symmetry and unlocks textures from degenerate TeX attributes, producing high-fidelity night vision with day-like visibility. Nevertheless, the library approach faces inherent obstacles. It rigidly asserts that each material possesses a fixed spectral emissivity, in stark contrast to the reality that emissivity is generally non-uniform, varying spatially and with viewing angle \cite{Jianrui2021, Gillespie1998}. Furthermore, a complete material library is prohibitive for unknown open scenes because of the enormous resources required for experimental calibration. Algorithmic library estimation is an alternative, but it requires the user’s prior information as input and is unstable due to the non-uniformity of emissivity. These obstacles undermine HADAR’s effectiveness, leaving universal anti-ghosting elusive even for the archetypal ghostly human faces, where non-uniform temperature and emissivity make texture recovery particularly challenging.

Here, we put forth a universal thermal anti-ghosting (TAG) framework for high-fidelity computational night vision (\fref{fig:fig1}a). We collect hyperspectral thermal imagery (\eg, by a Michelson interferometer) and then perform our proposed nonparametric spectral TeX decomposition (SLOT: Smoothness-structured Library-free Optimization for TeX; \fref{fig:fig1}b) that does not require material libraries. This novel computational thermal imaging framework naturally applies to unknown emissivity from open scenes, angularly non-uniform emissivity, or hybrid emissivity from mixed materials. We benchmark TAG on human faces (\fref{fig:fig1}c) and experimentally demonstrate vivid texture and expression recovery (\fref{fig:fig1}d) comparable to daylight optical imaging. This day-like clarity has been elusive so far due to unknown, non-uniform temperature and emissivity (Figs.~\ref{fig:fig2} and \ref{fig:fig3}), but proves crucial for facial and mood recognition as well as stereo reconstruction (\fref{fig:fig4}). TAG enables analysis of the role of material non-uniformity and reveals the effectiveness boundary of HADAR. We extensively test TAG on the DARPA Invisible Headlights Dataset \cite{yellin2024concurrent} (no material library available) and across both day and night (Figs.~\ref{fig:fig5} and \ref{fig:fig6}), demonstrating its universal advantage and robustness.

\section{Principles}
\label{sec:principle}

\subsection{HADAR rendering equation and the TeX degeneracy}
To illustrate our TAG framework, we first briefly recap the HADAR rendering equation \cite{Bao2023}. The total thermal radiance leaving an object $\alpha$ along $\tilde{z}$ direction is given by the rendering equation,
\begin{align}
\label{main_Eq}
S_{\alpha\nu}(\tilde{z})
&= e_{\alpha\nu}(\tilde{z})B_{\nu}(T_{\alpha}) \notag \\
&\quad + \int r_{\alpha\nu}(\tilde{z},\tilde{\rho})
  \bar{V}_{\alpha\beta}S_{\beta\nu}(\tilde{\rho})
  \,\mathrm{d}A_{\beta},
\end{align}
where the first term is the object's direct emission, and the second term is the ambient scattering from an environmental area $A_{\beta}$. Here, $\nu$ is the wavenumber. $e_{\alpha\nu}$ is the spectral emissivity, a unique material signature. In principle, $e_{\alpha\nu}$ has angular dependence, determined by the local surface normal $\tilde{A}_{\alpha}$ and the observing direction $\tilde{z}$. $B_{\nu}(T_{\alpha})$ is the blackbody radiation at temperature $T_{\alpha}$, governed by Planck’s law. $r_{\alpha\nu}(\tilde{z},\tilde{\rho})$ is the reflectance distribution function with light passing from $- \tilde{\rho}$ to $\tilde{z}$ direction. $\bar{V}_{\alpha\beta}=\frac{F(-\tilde{\rho}\cdot\tilde{A}_{\beta})F(\tilde{\rho}\cdot\tilde{A}_{\alpha})}{\pi\rho^{2}}$ is the differential view factor from $\beta$ to $\alpha$ satisfying $\int\bar{V}_{\alpha\beta}\mathrm{d}A_{\beta}=1$, $F(x)\equiv\max(0,x)$, and $\rho$ is the distance between objects $\alpha$ and $\beta$. 

To capture the underlying physics, we tackle the irreversible rendering equation in \eref{main_Eq} by evaluating the surface integral on each compact environmental object $\beta$ and replacing it with a single term, $r_{\alpha\nu}(\tilde{z})V_{\alpha\beta}S_{\beta\nu}$. This amounts to defining an averaged reflectivity and ambient radiance, $r_{\alpha\nu}(\tilde{z}) = \expv{r_{\alpha\nu}(\tilde{z},\tilde{\rho})}_{\tilde{\rho}}$, $S_{\beta\nu}=\expv{S_{\beta\nu}(\tilde{\rho})}_{\tilde{\rho}}$, and then defining the effective view factor $V_{\alpha\beta}=\int \eta_r\bar{V}_{\alpha\beta}\eta_S\mathrm{d}A_{\beta}$, where $\eta_r=r_{\alpha\nu}(\tilde{z},\tilde{\rho})/r_{\alpha\nu}(\tilde{z})$ and $\eta_S=S_{\beta\nu}(\tilde{\rho})/S_{\beta\nu}$. Approximately, we have $\sum_{\beta}V_{\alpha\beta}=1$. After applying Kirchhoff’s law, $r_{\alpha\nu}=1-e_{\alpha\nu}$, we arrive at the HADAR rendering equation
\begin{equation}
\label{main_function}
S_{\alpha\nu}=e_{\alpha\nu}B_{\nu}(T_{\alpha})+(1-e_{\alpha\nu})X_{\alpha\nu},
\end{equation}
with $X_{\alpha\nu}=\sum_{\beta\neq\alpha}V_{\alpha\beta}S_{\beta\nu}$. Here, we have absorbed the angular dependence $\tilde{z}$ and use the spatial index $\alpha$ to denote both spatial and angular material non-uniformity.

Experimental evidence shows that the texture X can be effectively truncated to the first two leading ambient contributions, typically originating from the sky and the average surrounding environment (conventionally denoted as the ground for simplicity) \cite{Bao2023}, that is, 
\begin{equation}\label{eq:X}
    X=VS_\mathrm{sky}+(1-V)S_\mathrm{g}.
\end{equation}
The `TeX degeneracy' refers to the fact that the thermal spectrum $S_\nu$ remains invariant if the object status shifts from the ground truth $\{T, e, V\}$ to the counterfactual $\{T', e', V'\}$, as long as
\begin{equation}
\begin{split}
    e'_{\nu} = &e_{\nu}\frac{B_{\nu}(T)-S_\mathrm{g}-V(S_\mathrm{sky}-S_\mathrm{g})}{B_{\nu}(T')-S_\mathrm{g}-V'(S_\mathrm{sky}-S_\mathrm{g})}\\
    &+ \frac{(V-V')(S_\mathrm{sky}-S_\mathrm{g})}{B_{\nu}(T')-S_\mathrm{g}-V'(S_\mathrm{sky}-S_\mathrm{g})}.
\end{split}
\end{equation}
This TeX degeneracy admits multiple counterfactual solutions beyond the ground truth and prevents accurate texture recovery for physics-agnostic approaches (\fref{fig:fig2}).

\subsection{SLOT with nonparametric emissivity}
Instead of introducing a rigid material library, here, we present SLOT (Smoothness-structured Library-free Optimization for TeX) to break the TeX symmetry (\fref{fig:fig1}b). SLOT models emissivity by a flexible cubic B-spline basis expansion and imposes a smoothness constraint. Explicitly, we have
\begin{equation}
    e(\nu) = \sum_{k=1}^{K}\beta_{k}\phi_{k}(\nu) \equiv \Phi(\nu)\boldsymbol{\beta}, \quad \nu\in[a,b],
    \label{eq:basis_expansion}
\end{equation}
where $\{\phi_{k}\}_{k=1}^{K}$ are cubic B-spline basis functions, \[
\Phi(\nu)=[\phi_1(\nu),\ldots,\phi_K(\nu)],
\qquad
\boldsymbol{\beta}=[\beta_{1},\ldots,\beta_{K}]^\top.
\] Cubic B-splines are used because they are smooth and locally supported, which yields stable numerics and allows local spectral structures to be expressed without global oscillations.

Robust TeX decomposition and texture recovery are guaranteed with joint physics constraints ($0<e<1$) and spectral-smoothness structures. For the latter, we penalize the discrete curvature of $\boldsymbol{\beta}$ via the second-order difference operator $D_{\beta}:$
\begin{equation}\label{eq:D_beta_def}
[D_\beta \beta]_j
 = \beta_j - 2\beta_{j+1} + \beta_{j+2},\qquad j = 1,\ldots,K-2.
\notag
\end{equation}
Explicitly, $D_\beta\in\mathbb R^{(K-2)\times K}$ is the banded matrix with rows
\begin{equation}
\begin{split}
&[D_\beta]_{j,j}=1,\; [D_\beta]_{j,j+1}=-2,\; [D_\beta]_{j,j+2}=1,\\
&\quad j=1,\ldots,K-2,
\notag
\end{split}
\end{equation}
and all other entries are zero. Finally, the penalty term becomes $\frac{\lambda}{2}\|D_\beta\beta\|_2^2$, where the regularization parameter $\lambda$ controls the trade-off between the data fidelity and the smoothness of the estimated emissivity. Importantly, conditioned on a given $\lambda$, this regularized formulation guarantees a unique globally optimal solution for TeX decomposition. See Secs.1 and 2 of the supplementary materials for more details of SLOT.

\subsection{Experimental Setup}
\label{subsec:experiment_setup}

The field thermal infrared hyperspectral data used in this study were acquired using a Fourier-transform thermal infrared (TIR) hyperspectral imaging spectrometer (Hypercam-LW, Telops Inc., Canada). The instrument employs a $320 \times 256$ mercury-cadmium-telluride (MCT) focal plane array, providing an instantaneous field of view (IFOV) of 0.35 mrad. Taking advantage of the Fourier-transform spectrometer (FTS) architecture, the effective spectral response spanned the $870-1269 \text{ cm}^{-1}$ range. The actual spectral resolution was set to $6 \text{ cm}^{-1}$, though the system is capable of up to $0.25 \text{ cm}^{-1}$.

The data acquisition was conducted outdoors in an open environment in Wuhan, China, on January 6 and November 30, 2025. During the experiments, the ambient air temperature was approximately $17^\circ\text{C}$, the relative humidity was below 27\%, and the weather conditions were clear and cloud-free. The targets (human faces) were oriented toward a direction free of significant anthropogenic thermal radiation sources. While airborne thermal hyperspectral imaging typically requires complex atmospheric compensation and noise-resilient algorithms to retrieve temperature and emissivity \cite{WANG_2026, WANG_2024}, our close-range experiment deliberately avoids these transmission disturbances. To minimize atmospheric transmittance effects along the optical path, the distance between the target and the sensor was fixed at the minimum focusing distance of the Hypercam-LW, ensuring stable imaging geometry. To accurately formulate the texture $X$, additional sky-looking and ground-looking hyperspectral measurements were acquired along the identical viewing direction immediately after capturing the targets.

\begin{figure*}[t!]
    \centering
    \scalebox{1}{\includegraphics{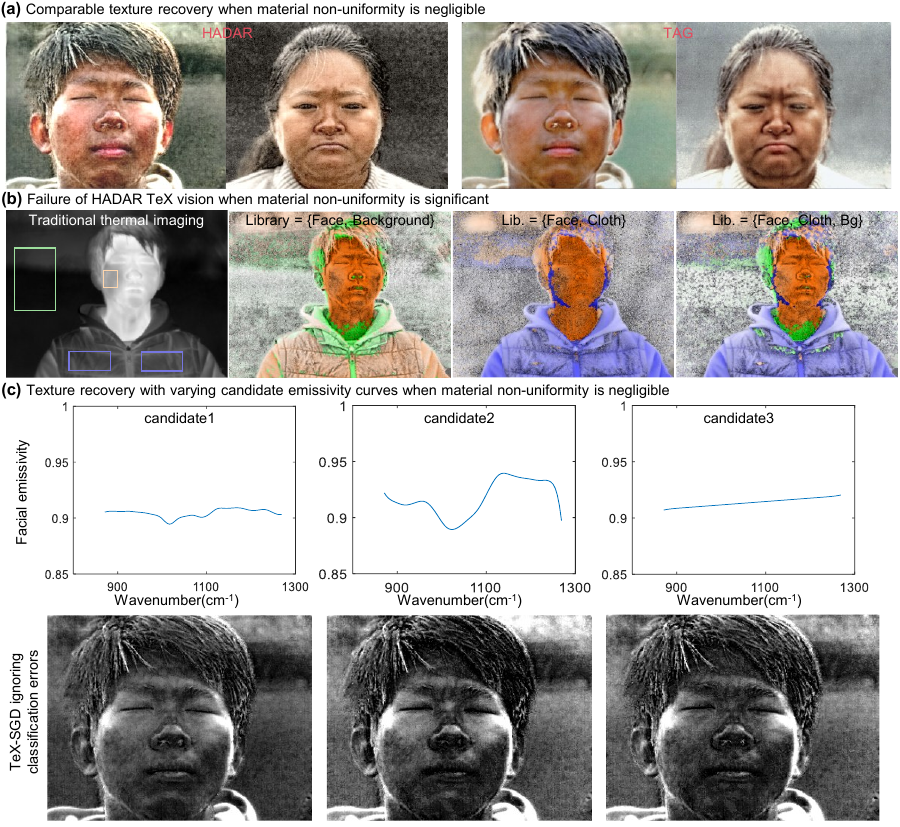}}
    \caption{Shedding light on the effectiveness boundary of HADAR. a, Comparable texture recovery between HADAR and TAG when intra-class material non-uniformity is smaller (negligible) than the inter-class emissivity contrast. b, Failure of HADAR TeX vision when intra-class material non-uniformity is larger (significant) than the inter-class emissivity contrast. c, HADAR is robust to emissivity/library inaccuracy when material non-uniformity is negligible. The leftmost panel uses the baseline candidate emissivity estimated by the HADAR approach. The middle and right panels use candidate curves from TAG controlled by the regularization parameter ($\lambda = 10^{-3}, 10^6$).}
    \label{fig:fig3} 
\end{figure*}
\begin{figure*}[t!]
    \centering
    \scalebox{1}{\includegraphics{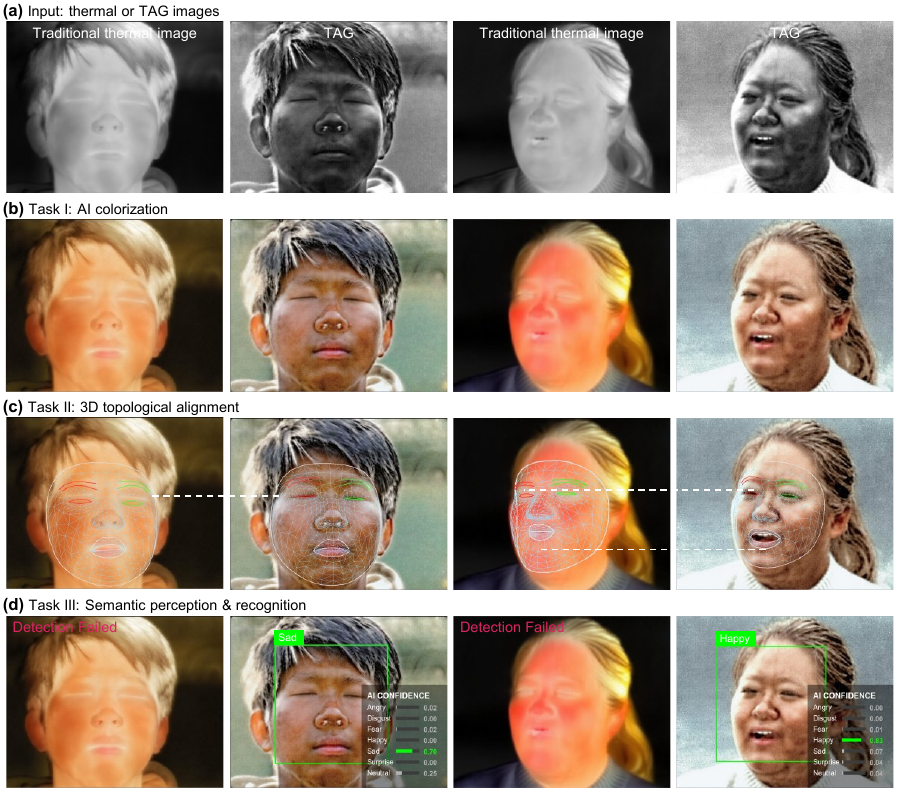}}
    \caption{Zero-shot cross-modal machine perception enabled by TAG. a, Input data comparison showing ghosting traditional thermal imaging and TAG with vivid textures. b, Task I: AI colorization. With textures, TAG perfectly accommodates color mapping and preserves spatial patterns. Without textures, traditional thermal images produce blurry color artifacts. c, Task II: 3D topological alignment. TAG recovers textures to flawlessly anchor a 468-point facial mesh, whereas alignment completely fails or drifts on traditional thermal images. d, Task III: Semantic perception and recognition. Standard RGB-trained vision engines successfully localize faces and predict nuanced human emotions (\ie, Sad, Happy) on TAG images, but suffer catastrophic breakdowns on thermal images.}
    \label{fig:fig4} 
\end{figure*}

\begin{figure}[h]
    \centering
    \includegraphics[width=\columnwidth]{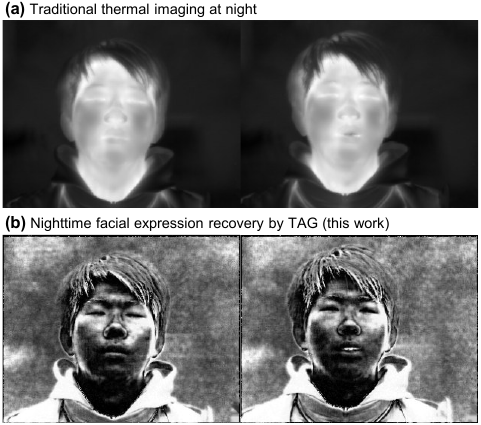}
    \captionsetup{width=\columnwidth}
    \caption{Passive recovery of facial texture and expression with high-fidelity night vision in pitch darkness. a, Ghosting thermal imaging at night. b, Nighttime facial expression recovery by TAG. Left: neutral expression with mouth and eyes closed. Right: Smiling with mouth and eyes open.}
    \label{fig:fig5} 
\end{figure}

\begin{figure*}[htb]
    \centering
    \scalebox{1}{\includegraphics{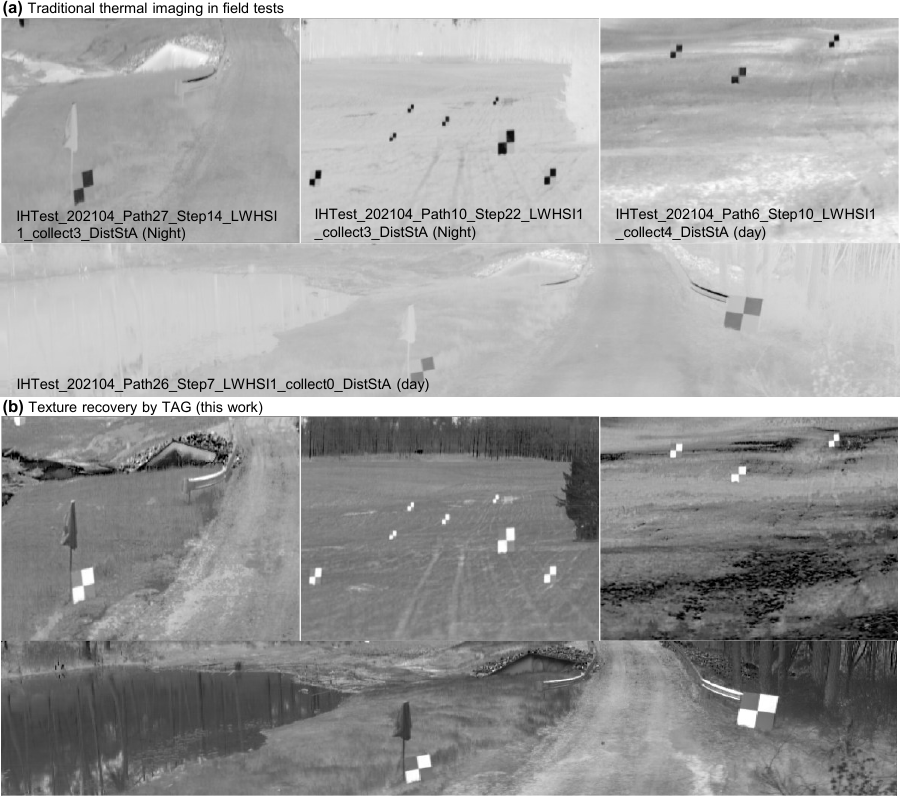}}
    \caption{Robustness of TAG in complex open-world field tests. a, Traditional thermal imaging from the DARPA Invisible Headlights dataset, where the ghosting effect blurs critical details. b, TAG consistently recovers great textures of unknown materials across vast scenes without requiring prior material libraries (surpassing HADAR), both during the day and at night.}
    \label{fig:fig6} 
\end{figure*}

\section{Results}
\label{sec:results}

\subsection{Texture and expression recovery for ghostly human faces}
The ghosting effect is most famously observed in thermal imaging of human faces with highly non-uniform temperature and emissivity (\fref{fig:fig1}c). \fref{fig:fig2}b (left) shows the failure of texture recovery by traditional image post-processing: CLAHE enhances the visual contrast but cannot completely remove the influence of non-uniform temperature and emissivity in the presence of the TeX degeneracy. With our TAG, \fref{fig:fig1}d (also see the right panel of \fref{fig:fig1}b) demonstrates vivid texture recovery of the archetypal ghostly human faces across both male and female, with various expressions (see Sec.5 of the supplementary material for colorization details). We emphasize that the visibility of our recovered facial textures and expressions matches that of daylight optical imaging (see Fig.S3 in the supplementary materials), a result that was previously elusive.

Most significantly, \fref{fig:fig1}a shows the decomposed temperature and spectral emissivity curves, both of which exhibit spatial non-uniformity. Particularly, in areas 1 (red solid box) and 2 (cyan dashed box) on the face, the corresponding mean (curve) and standard deviation (shade) of the spectral emissivity are drawn, correctly confirming the non-uniformity around different areas of the same material. Note that emissivity non-uniformity may arise from multiple effects, including the angle dependence of Fresnel's reflection, surface roughness, non-uniformly oily skin, and so on.

\subsection{Applicability of HADAR material classification}
Our TAG framework allows continuous control of spectral emissivity by tuning the smoothness regularization parameter $\lambda$. This enables testing the sensitivity of TAG with respect to regularization, the consistency between TAG and HADAR, and the influence of material non-uniformity on HADAR material classification.

Firstly, for the face scene in \fref{fig:fig3}c, we constructed a material library with two effective materials, the face and the background, and evaluated HADAR textures (left panel). Their emissivity curves were estimated from the data by traditional temperature-emissivity separation (TES; background emissivity not shown). Secondly, we applied TAG on the same scene and observed that regularization with $\lambda=0.22$ reproduces the TES and HADAR textures, whereas other $\lambda$ values yield different decomposed emissivities (right two panels). Thirdly, we replaced the facial emissivity with one of the latter two emissivity curves decomposed by TAG, and then used the material library with the HADAR approach for TeX decomposition. Interestingly, all three experiments yielded fine, comparable textures, demonstrating the robustness of the HADAR approach to moderate emissivity variations and library inaccuracies. This observation implies that the `ground-truth' material library is, in fact, non-essential to HADAR to overcome the ghosting effect. As long as most pixels can be correctly classified into the desired materials, the material library works, and the material non-uniformity has nearly negligible influence — the origin of the algorithmically estimated library's success in HADAR. We emphasize that HADAR's performance depends on the explicit, estimated material library and hence becomes user-dependent. Once optimized by estimating emissivity in different regions through trial and error, HADAR achieves textures (\fref{fig:fig3}a) comparable to TAG.

However, when a material’s internal non-uniformity exceeds the contrast between distinct materials, material classification always fails regardless of how the library is constructed, resulting in texture patches with sharp boundaries inside the same material region. \fref{fig:fig3}b demonstrates the failure of the cutting-edge HADAR TeX vision, with the library estimated in corresponding boxed areas. We emphasize that material classification errors are inevitable due to the significant non-uniformity of emissivity, even though the errors vary across different library configurations. This delineates the effectiveness boundary for the existing HADAR approach. On the contrary, TAG consistently recovers vivid textures as shown in \fref{fig:fig1}b, despite the material non-uniformity.

\subsection{Objective texture quantification in machine perception tasks}
To rigorously quantify the fidelity of the recovered textures, we evaluated the results using both low-level physical metrics and a high-level machine perception pipeline. For a fair comparison, we evaluated the following established image quality metrics: Information Entropy (EN), Average Gradient (AG), Spatial Frequency (SF), and Standard Deviation (SD) (see Sec.6 of the supplementary materials for the specific definitions), for the foreground human region in \fref{fig:fig4}a. Tab.~\ref{tab:quantification} consistently shows that TAG significantly outperforms traditional thermal imaging across all metrics. Note that HADAR is not included in the comparison, as it requires user input and does not universally apply in the presence of material non-uniformity.
\begin{table}
\centering
\caption{Texture quantification comparing two universal modalities, the traditional thermal imaging (IR) and our TAG, in the human region in \fref{fig:fig4}a.}
\label{tab:quantification}
\resizebox{\columnwidth}{!}{
\begin{tabular}{llcccc}
\toprule
\multicolumn{2}{c}{Metrics} & EN $\uparrow$ & AG $\uparrow$ & SF $\uparrow$ & SD $\uparrow$ \\
\midrule
\multirow{2}{*}{Male} & IR & 7.349 & 0.476 & 1.331 & 51.491 \\
 & \textbf{TAG (Ours)} & \cellcolor{red!10}\textbf{7.491} & \cellcolor{red!10}\textbf{1.916} & \cellcolor{red!10}\textbf{5.933} & \cellcolor{red!10}\textbf{53.079} \\
\midrule
\multirow{2}{*}{Female} & IR & 6.761 & 1.329 & 3.408 & 50.523 \\
 & \textbf{TAG (Ours)} & \cellcolor{red!10}\textbf{7.412} & \cellcolor{red!10}\textbf{4.157} & \cellcolor{red!10}\textbf{8.363} & \cellcolor{red!10}\textbf{52.004} \\
\bottomrule
\end{tabular}
}
\end{table}

Building upon this rich physical foundation, we evaluated AI colorization (Task I; see the supplementary materials). As shown in \fref{fig:fig4}b, a basic colorization process naturally adapts to the TAG images, successfully mapping realistic tones while strictly preserving spatial patterns. Conversely, applying exactly the same algorithm to traditional thermal images yields blurry color blobs, revealing the irrecoverable loss of textures in the original inputs.

Subsequently, we performed 3D topological alignment (Task II) and semantic perception (Task III) on these colorized outputs (see Sec.4 in the supplementary materials for tests on the grayscale inputs with consistent conclusions). For Task II, the Google MediaPipe framework \cite{kartynnik2019} flawlessly anchors a 468-point 3D facial mesh onto the TAG images, whereas it completely fails to locate geometric anchors on the traditional counterparts. For Task III, utilizing an MTCNN detector \cite{Zhang_2016} coupled with the \texttt{fer} classification library \cite{arriaga2017}, the vision engine seamlessly localizes faces and predicts nuanced human emotions (\eg, `Happy' and `Sad') with high confidence scores on the TAG images. Not surprisingly, traditional thermal images lead to catastrophic failures and missed detections.

\subsection{Turning night into day}

We present our striking demonstration of passive recovery of facial texture and expression in pitch darkness with TAG (\fref{fig:fig5}), advancing the frontier of `turning night into day' \cite{Pile2023}. While RGB imaging is nearly completely black (not shown), thermal imaging captures the contours but with poor detail, so facial expressions are barely visible. In stark contrast, TAG overcomes ghosting and restores vivid textures and facial expressions, as if it were day, a result that was elusive so far. This result can lead to high-fidelity and universal computational night vision, enabling seamless machine perception across day and night. Note that TAG textures are generated by sky thermal illumination through the view factor $V$ in \eref{eq:X}. At night, sky radiance is substantially weaker than during the day due to lower atmospheric temperatures, resulting in a lower signal-to-noise (SNR) ratio in \fref{fig:fig5} than in \fref{fig:fig1}d. We emphasize that the presented night experiments were deliberately conducted on a winter night, representing the most extreme condition where the SNR approaches its annual minimum. The fact that TAG successfully recovers clear geometric textures under such low SNR conditions rigorously guarantees the year-round operational reliability of our framework. Additionally, because the ground has a lower reflectivity (around 0.05) in the thermal infrared than in the visible-light range, the face appears brighter in up-oriented areas (\eg, the temple and the upper cheek) and darker in down-oriented areas (\eg, the chin and the lower cheek), compared to conventional grayscale optical imaging.

\subsection{Robustness test on the DARPA Invisible Headlights dataset}

We conducted extensive robustness tests on the large-scale DARPA Invisible Headlights dataset \cite{yellin2024concurrent} to demonstrate that TAG generalizes beyond human targets (see \fref{fig:fig6} for representative examples; the majority of test results are not shown for brevity). \fref{fig:fig6}a shows ghosting thermal images, where complex field scenes are blurry and almost indistinguishable regardless of whether the data was collected during the day or at night. With our TAG, \fref{fig:fig6}b demonstrates vivid geometric texture recovery across vast scenes, both during the day and at night.

Most importantly, TAG successfully resolves great details of diverse unknown materials (\eg, roads, water, trees, calibration boards, etc.) without relying on any prior information, surpassing HADAR, which requires a pre-calibrated material library. TAG thus bypasses the out-of-distribution (OOD) failures that plague material classification approaches. Note that in open environments, OOD errors inevitably arise due to the infinite variety of natural materials, spatially varying surface roughness, and complex environmental scattering, which prevent simple calibration within a finite library.

\section{Conclusion}
\label{sec:discussion}

In summary, we have introduced a computational thermal imaging framework, TAG, that addresses material non-uniformity, thereby enabling universal ghosting mitigation and high-fidelity night vision. TAG collects hyperspectral thermal imagery and proposes SLOT, which uses smoothness structures in spectral emissivity to break TeX symmetry. We demonstrated thermal anti-ghosting with day-like visibility across diverse scenes and across day and night. In particular, we tackled the thus far elusive, archetypal ghosting phenomenon in thermal imaging of human faces, where non-uniform temperature and emissivity make anti-ghosting extremely challenging. Our unprecedented facial texture and expression recovery set the benchmark and demonstrated improved performance on machine perception tasks, including 3D topological alignment and semantic recognition. This leads to a next-generation HADAR, without any priors, for seamless machine perception across day and night. Our results strongly urge the development of real-time, portable, and cost-effective photonic chips with integrated spectral discrimination (\eg, via metasurfaces) for next-generation night vision. We believe our work will reshape how machines and humans perceive the invisible world and lead to the ubiquitous adoption of high-fidelity night vision in the era of AI.

\section*{Disclosures} 
The authors declare no conflicts of interest.

\section*{Data Availability} 
Data underlying the results presented in this paper are not publicly available at this time but may be obtained from the authors upon reasonable request.

\section*{Supplemental Document}
See Supplement 1 for supporting content.

\bibliographystyle{unsrt}
\bibliography{reference}

@article{lawson1956implications,
  title={Implications of surface temperatures in the diagnosis of breast cancer},
  author={Lawson, Ray},
  journal={Can. Med. Assoc. J.},
  volume={75},
  number={4},
  pages={309},
  year={1956},
  url={https://pmc.ncbi.nlm.nih.gov/articles/PMC1824571/}
}

@article{schmidt2014thermal,
  title={Thermal camera and method for eliminating ghosting effects of hot-target thermal images},
  author={Schmidt, Matthew F and Mumaw, David T},
  year={2014},
  month=jul # "~3",
  journal={Google Patents},
  pages={US Patent App. 13/733,647},
  url = {https://www.freepatentsonline.com/y2014/0184805.html}
}

@INPROCEEDINGS{Rafael2021,
  author={Rivadeneira, Rafael E. and Sappa, Angel D. and Vintimilla, Boris X. and Nathan, Sabari and Kansal, Priya and Mehri, Armin and Behjati Ardakani, Parichehr and Dalal, Anurag and Akula, Aparna and Sharma, Darshika and Pandey, Shashwat and Kumar, Basant and Yao, Jiaxin and Wu, Rongyuan and Feng, Kai and Li, Ning and Zhao, Yongqiang and Patel, Heena and Chudasama, Vishal and Prajapati, Kalpesh and Sarvaiya, Anjali and Upla, Kishor P. and Raja, Kiran and Ramachandra, Raghavendra and Busch, Christoph and Almasri, Feras and Vandamme, Thomas and Debeir, Olivier and Gutierrez, Nolan B. and Nguyen, Quan H. and Beksi, William J.},
  booktitle={2021 IEEE/CVF Conference on Computer Vision and Pattern Recognition Workshops (CVPRW)}, 
  title={Thermal Image Super-Resolution Challenge - PBVS 2021}, 
  year={2021},
  volume={},
  number={},
  pages={4354-4362},
  doi={10.1109/CVPRW53098.2021.00492}
}

@InProceedings{Treible_2017_CVPR,
author = {Treible, Wayne and Saponaro, Philip and Sorensen, Scott and Kolagunda, Abhishek and O'Neal, Michael and Phelan, Brian and Sherbondy, Kelly and Kambhamettu, Chandra},
title = {CATS: A Color and Thermal Stereo Benchmark},
booktitle = {Proceedings of the IEEE Conference on Computer Vision and Pattern Recognition (CVPR)},
month = {July},
year = {2017},
doi = {10.1109/CVPR.2017.22},
}

@Article{Gade2014,
  author        = {Gade, Rikke and Moeslund, Thomas B.},
  title         = {Thermal cameras and applications: a survey},
  journal       = {Mach. Vis. Appl.},
  year          = {2014},
  volume        = {25},
  number        = {1},
  pages         = {245--262},
  month         = jan,
  issn          = {1432-1769},
  doi           = {10.1007/s00138-013-0570-5},
}

@Article{Bao2023,
  author  = {Bao, Fanglin and Wang, Xueji and Sureshbabu, Shree Hari and Sreekumar, Gautam and Yang, Liping and Aggarwal, Vaneet and Boddeti, Vishnu N. and Jacob, Zubin},
  journal = {Nature},
  title   = {Heat-assisted detection and ranging},
  year    = {2023},
  issn    = {1476-4687},
  number  = {7971},
  pages   = {743--748},
  volume  = {619},
  doi     = {10.1038/s41586-023-06174-6}
}

@article{Bao2024,
author = {Fanglin Bao and Shubhankar Jape and Andrew Schramka and Junjie Wang and Tim E. McGraw and Zubin Jacob},
journal = {Opt. Express},
number = {3},
pages = {3852--3865},
publisher = {Optica Publishing Group},
title = {Why thermal images are blurry},
volume = {32},
month = {Jan},
year = {2024},
doi     = {10.1364/OE.506634}
}

@article{Herschel1800,
  title={XIII. Investigation of the powers of the prismatic colours to heat and illuminate objects; with remarks, that prove the different refrangibility of radiant heat. To which is added, an inquiry into the method of viewing the sun advantageously, with telescopes of large apertures and high magnifying powers},
  author={Herschel, William},
  journal={Philos. Trans. R. Soc. London},
  number={90},
  pages={255--283},
  year={1800},
  publisher={The Royal Society London},
  doi = {10.1098/rspl.1800.0012},
  url = {https://doi.org/10.1098/rspl.1800.0012},
}

@inproceedings{yellin2024concurrent, 
title={Concurrent Band Selection and Traversability Estimation From Long-Wave Hyperspectral Imagery in Off-Road Settings}, 
author={Yellin, Florence and McCloskey, Scott and Hill, Cole and Smith, Eric and Clipp, Brian}, 
booktitle={Proceedings of the IEEE/CVF Winter Conference on Applications of Computer Vision}, 
pages={7483--7492}, 
year={2024},
doi={10.1109/WACV57701.2024.00731}
}

@article{Soundrapandiyan2022,
Author = {Soundrapandiyan, Rajkumar and Satapathy, Suresh Chandra and Mouli, P. V.
   S. S. R. Chandra and Nhu, Nguyen Gia},
Title = {A comprehensive survey on image enhancement techniques with special
   emphasis on infrared images},
Journal = {Multim. Tools Appl.},
Year = {2022},
Volume = {81},
Number = {7},
Pages = {9045-9077},
Month = {MAR},
DOI = {10.1007/s11042-021-11250-y},
}

@ARTICLE{TEFormer2025,
  author={Wu, Zhe and Li, Yunxin and Zhang, Runmin and Cao, Si-Yuan and Ying, Jiacheng and Zhang, Xiaohan and Bai, Xiaokai and Chen, Shujie and Yang, Bailin and Shen, Hui-Liang},
  journal={IEEE Trans. Geosci. Remote Sens.}, 
  title={TEFormer: Thermal Infrared Image Enhancement by Preserving Spatial Consistency and Details}, 
  year={2025},
  volume={63},
  number={},
  pages={1-14},
  doi={10.1109/TGRS.2025.3611503}
}

@article{XU2025106114,
title = {A multi-scale perception network for infrared thermal image super-resolution in welding},
journal = {Infrared Phys. Technol.},
volume = {151},
pages = {106114},
year = {2025},
issn = {1350-4495},
doi = {10.1016/j.infrared.2025.106114},
author = {Qingpo Xu and Haitao Liu and Jiameng Gao and Yabin Ding and Juliang Xiao and Guangxi Li},
}

@Article{Pile2023,
  author  = {Pile, David},
  journal = {Nat. Photonics},
  title   = {Turning night into day},
  year    = {2023},
  issn    = {1749-4893},
  number  = {10},
  pages   = {843--843},
  volume  = {17},
  doi     = {10.1038/s41566-023-01297-8},
  refid   = {Pile2023}
}

@Article{Jianrui2021,
AUTHOR = {Hu, Jianrui and Liu, Zhanqiang and Zhao, Jinfu and Wang, Bing and Song, Qinghua},
TITLE = {Theoretical Modeling and Analysis of Directional Spectrum Emissivity and Its Pattern for Random Rough Surfaces with a Matrix Method},
JOURNAL = {Symmetry},
VOLUME = {13},
YEAR = {2021},
NUMBER = {9},
ARTICLE-NUMBER = {1733},
ISSN = {2073-8994},
DOI = {10.3390/sym13091733}
}

@Article{Alejandro2016,
AUTHOR = {González, Alejandro and Fang, Zhijie and Socarras, Yainuvis and Serrat, Joan and Vázquez, David and Xu, Jiaolong and López, Antonio M.},
TITLE = {Pedestrian Detection at Day/Night Time with Visible and FIR Cameras: A Comparison},
JOURNAL = {Sensors},
VOLUME = {16},
YEAR = {2016},
NUMBER = {6},
ARTICLE-NUMBER = {820},
PubMedID = {27271635},
ISSN = {1424-8220},
DOI = {10.3390/s16060820}
}

@inproceedings{Xinyu2025,
author = {Hou, Xinyu and Beuchert, Jonas and Shin, Sangyun and Markham, Andrew and Trigoni, Niki},
title = {Thermal-to-RGB Video Translation for Wildlife Monitoring: Enhancing Low-Resolution Thermal Imagery with Large Diffusion Models},
year = {2025},
isbn = {9798400719820},
publisher = {Association for Computing Machinery},
address = {New York, NY, USA},
doi = {10.1145/3737905.3769285},
booktitle = {Proceedings of the 2025 ACM International Workshop on Thermal Sensing and Computing},
pages = {13–18},
numpages = {6},
location = {Hong Kong, China},
series = {HotSense'25}
}

@Article{Gillespie1998,
  author        = {A. {Gillespie} and S. {Rokugawa} and T. {Matsunaga} and J. S. {Cothern} and S. {Hook} and A. B. {Kahle}},
  title         = {A temperature and emissivity separation algorithm for Advanced Spaceborne Thermal Emission and Reflection Radiometer (ASTER) images},
  journal       = {IEEE Trans. Geosci. Remote Sens.},
  year          = {1998},
  volume        = {36},
  number        = {4},
  pages         = {1113-1126},
  month         = {July},
  issn          = {1558-0644},
  doi           = {10.1109/36.700995},
}

@article{ROGALSKI2011136,
title = {Recent progress in infrared detector technologies},
journal = {Infrared Physics \& Technology},
volume = {54},
number = {3},
pages = {136-154},
year = {2011},
note = {Proceedings of the International Conference on Quantum Structure Infrared Photodetector (QSIP) 2010},
issn = {1350-4495},
doi = {https://doi.org/10.1016/j.infrared.2010.12.003},
url = {https://www.sciencedirect.com/science/article/pii/S1350449510001040},
author = {A. Rogalski},
}

@INPROCEEDINGS{Hwang2015,
  author={Hwang, Soonmin and Park, Jaesik and Kim, Namil and Choi, Yukyung and Kweon, In So},
  booktitle={2015 IEEE Conference on Computer Vision and Pattern Recognition (CVPR)}, 
  title={Multispectral pedestrian detection: Benchmark dataset and baseline}, 
  year={2015},
  volume={},
  number={},
  pages={1037-1045},
  doi={10.1109/CVPR.2015.7298706}}

@ARTICLE{Sun2019,
  author={Sun, Yuxiang and Zuo, Weixun and Liu, Ming},
  journal={IEEE Robotics and Automation Letters}, 
  title={RTFNet: RGB-Thermal Fusion Network for Semantic Segmentation of Urban Scenes}, 
  year={2019},
  volume={4},
  number={3},
  pages={2576-2583},
  doi={10.1109/LRA.2019.2904733}}

@article{GUAN2019,
title = {Fusion of multispectral data through illumination-aware deep neural networks for pedestrian detection},
journal = {Information Fusion},
volume = {50},
pages = {148-157},
year = {2019},
issn = {1566-2535},
doi = {https://doi.org/10.1016/j.inffus.2018.11.017},
url = {https://www.sciencedirect.com/science/article/pii/S1566253517308138},
author = {Dayan Guan and Yanpeng Cao and Jiangxin Yang and Yanlong Cao and Michael Ying Yang},
}

@inproceedings{Zhou2020,
author = {Zhou, Kailai and Chen, Linsen and Cao, Xun},
title = {Improving Multispectral Pedestrian Detection by Addressing Modality Imbalance Problems},
year = {2020},
isbn = {978-3-030-58522-8},
publisher = {Springer-Verlag},
address = {Berlin, Heidelberg},
url = {https://doi.org/10.1007/978-3-030-58523-5_46},
doi = {10.1007/978-3-030-58523-5_46},
booktitle = {Computer Vision – ECCV 2020: 16th European Conference, Glasgow, UK, August 23–28, 2020, Proceedings, Part XVIII},
pages = {787–803},
numpages = {17},
location = {Glasgow, United Kingdom}
}

@ARTICLE{Chio2018,
  author={Choi, Yukyung and Kim, Namil and Hwang, Soonmin and Park, Kibaek and Yoon, Jae Shin and An, Kyounghwan and Kweon, In So},
  journal={IEEE Transactions on Intelligent Transportation Systems}, 
  title={KAIST Multi-Spectral Day/Night Data Set for Autonomous and Assisted Driving}, 
  year={2018},
  volume={19},
  number={3},
  pages={934-948},
  doi={10.1109/TITS.2018.2791533}}

@article{Khattak2020,
author = {Khattak, Shehryar and Papachristos, Christos and Alexis, Kostas},
title = {Keyframe-based thermal–inertial odometry},
journal = {Journal of Field Robotics},
volume = {37},
number = {4},
pages = {552-579},
doi = {https://doi.org/10.1002/rob.21932},
url = {https://onlinelibrary.wiley.com/doi/abs/10.1002/rob.21932},
eprint = {https://onlinelibrary.wiley.com/doi/pdf/10.1002/rob.21932},
year = {2020}
}

@INPROCEEDINGS{Riggan2018,
  author={Riggan, Benjamin S. and Short, Nathaniel J. and Hu, Shuowen},
  booktitle={2018 IEEE Winter Conference on Applications of Computer Vision (WACV)}, 
  title={Thermal to Visible Synthesis of Face Images Using Multiple Regions}, 
  year={2018},
  volume={},
  number={},
  pages={30-38},
  doi={10.1109/WACV.2018.00010}}

@INPROCEEDINGS{Ha2017,
  author={Ha, Qishen and Watanabe, Kohei and Karasawa, Takumi and Ushiku, Yoshitaka and Harada, Tatsuya},
  booktitle={2017 IEEE/RSJ International Conference on Intelligent Robots and Systems (IROS)}, 
  title={MFNet: Towards real-time semantic segmentation for autonomous vehicles with multi-spectral scenes}, 
  year={2017},
  volume={},
  number={},
  pages={5108-5115},
  doi={10.1109/IROS.2017.8206396}}

@INPROCEEDINGS{Shivakumar2020,
  author={Shivakumar, Shreyas S. and Rodrigues, Neil and Zhou, Alex and Miller, Ian D. and Kumar, Vijay and Taylor, Camillo J.},
  booktitle={2020 IEEE International Conference on Robotics and Automation (ICRA)}, 
  title={PST900: RGB-Thermal Calibration, Dataset and Segmentation Network}, 
  year={2020},
  volume={},
  number={},
  pages={9441-9447},
  doi={10.1109/ICRA40945.2020.9196831}}

@InProceedings{Yash2021,
author="Khare, Yash
and Pavithran, Vipin",
editor="Singh, Kehar
and Gupta, A. K.
and Khare, Sudhir
and Dixit, Nimish
and Pant, Kamal",
title="Infrared Image Enhancement Using Convolution Matrices",
booktitle="ICOL-2019",
year="2021",
publisher="Springer Singapore",
address="Singapore",
pages="337--340",
isbn="978-981-15-9259-1"
}

@article{Dhal_2021,
	title = {Histogram {Equalization} {Variants} as {Optimization} {Problems}: {A} {Review}},
	volume = {28},
	issn = {1886-1784},
	url = {https://doi.org/10.1007/s11831-020-09425-1},
	doi = {10.1007/s11831-020-09425-1},
	number = {3},
	journal = {Archives of Computational Methods in Engineering},
	author = {Dhal, Krishna Gopal and Das, Arunita and Ray, Swarnajit and Gálvez, Jorge and Das, Sanjoy},
	month = may,
	year = {2021},
	pages = {1471--1496},
}

@article{LI_2018,
title = {An improved contrast enhancement algorithm for infrared images based on adaptive double plateaus histogram equalization},
journal = {Infrared Physics \& Technology},
volume = {90},
pages = {164-174},
year = {2018},
issn = {1350-4495},
doi = {https://doi.org/10.1016/j.infrared.2018.03.010},
url = {https://www.sciencedirect.com/science/article/pii/S1350449517306503},
author = {Shuo Li and Weiqi Jin and Li Li and Yiyang Li},
}

@Article{Li_2023,
AUTHOR = {Li, Huaizhou and Wang, Shuaijun and Bai, Zhenpeng and Wang, Hong and Li, Sen and Wen, Shupei},
TITLE = {Research on 3D Reconstruction of Binocular Vision Based on Thermal Infrared},
JOURNAL = {Sensors},
VOLUME = {23},
YEAR = {2023},
NUMBER = {17},
ARTICLE-NUMBER = {7372},
URL = {https://www.mdpi.com/1424-8220/23/17/7372},
PubMedID = {37687828},
ISSN = {1424-8220},
DOI = {10.3390/s23177372}
}

@InProceedings{Bouhlel_2023,
author="Bouhlel, Fatma
and Mliki, Hazar
and Lagha, Rayen
and Hammami, Mohamed",
editor="Abraham, Ajith
and Pllana, Sabri
and Casalino, Gabriella
and Ma, Kun
and Bajaj, Anu",
title="TIR-GAN: Thermal Images Restoration Using Generative Adversarial Network",
booktitle="Intelligent Systems Design and Applications",
year="2023",
publisher="Springer Nature Switzerland",
address="Cham",
pages="428--437",
isbn="978-3-031-35507-3"
}

@inproceedings{Pang_2023,
author = {Zhongxiang Pang and Guihua Liu and Guosheng Li and Jian Gong and Chunmei Chen and Chao Yao},
title = {{A two-stream deep neural network for infrared image enhancement}},
volume = {12563},
booktitle = {AOPC 2022: AI in Optics and Photonics},
editor = {Chao Zuo},
organization = {International Society for Optics and Photonics},
publisher = {SPIE},
pages = {1256302},
year = {2023},
doi = {10.1117/12.2642633},
URL = {https://doi.org/10.1117/12.2642633}
}

@article{KUANG_2019,
title = {Single infrared image enhancement using a deep convolutional neural network},
journal = {Neurocomputing},
volume = {332},
pages = {119-128},
year = {2019},
issn = {0925-2312},
doi = {https://doi.org/10.1016/j.neucom.2018.11.081},
url = {https://www.sciencedirect.com/science/article/pii/S0925231218314905},
author = {Xiaodong Kuang and Xiubao Sui and Yuan Liu and Qian Chen and Guohua Gu},
}

@ARTICLE{Lee_2017,
  author={Lee, Kyungjae and Lee, Junhyeop and Lee, Joosung and Hwang, Sangwon and Lee, Sangyoun},
  journal={IEEE Access}, 
  title={Brightness-Based Convolutional Neural Network for Thermal Image Enhancement}, 
  year={2017},
  volume={5},
  number={},
  pages={26867-26879},
  doi={10.1109/ACCESS.2017.2769687}}

@inproceedings{Hou2025,
author = {Hou, Xinyu and Beuchert, Jonas and Shin, Sangyun and Markham, Andrew and Trigoni, Niki},
title = {Thermal-to-RGB Video Translation for Wildlife Monitoring: Enhancing Low-Resolution Thermal Imagery with Large Diffusion Models},
year = {2025},
isbn = {9798400719820},
publisher = {Association for Computing Machinery},
address = {New York, NY, USA},
url = {https://doi.org/10.1145/3737905.3769285},
doi = {10.1145/3737905.3769285},
booktitle = {Proceedings of the 2025 ACM International Workshop on Thermal Sensing and Computing},
pages = {13–18},
numpages = {6},
location = {Hong Kong, China},
series = {HotSense '25}
}

@inproceedings{Wen2025,
author = {Wen, Xiangyu and Fang, Guangchi and Yang, Bo and Wang, Bing},
title = {Geometry Aware 3D Multiview Thermal Reconstruction with Emissive Residual Decomposition Gaussian Splatting},
year = {2025},
isbn = {9798400719820},
publisher = {Association for Computing Machinery},
address = {New York, NY, USA},
url = {https://doi.org/10.1145/3737905.3769282},
doi = {10.1145/3737905.3769282},
booktitle = {Proceedings of the 2025 ACM International Workshop on Thermal Sensing and Computing},
pages = {7–12},
numpages = {6},
location = {Hong Kong, China},
series = {HotSense '25}
}

@inproceedings{Chen2025,
author = {Chen, Yuhan and Song, Jingwei and Zhang, Xie and Zhang, Jianqi and Wu, Chenshu},
title = {ThermalEye: Fully Passive Eye Blink Detection on Smart Glasses via Low-Cost Thermal Sensing},
year = {2025},
isbn = {9798400719820},
publisher = {Association for Computing Machinery},
address = {New York, NY, USA},
url = {https://doi.org/10.1145/3737905.3769280},
doi = {10.1145/3737905.3769280},
booktitle = {Proceedings of the 2025 ACM International Workshop on Thermal Sensing and Computing},
pages = {34–39},
numpages = {6},
location = {Hong Kong, China},
series = {HotSense '25}
}

@article{beaudry2012intuitive,
  title={An intuitive proof of the data processing inequality},
  author={Beaudry, Normand J and Renner, Renato},
  journal={Quantum Information \& Computation},
  volume={12},
  number={5-6},
  pages={432--441},
  year={2012},
  publisher={Rinton Press, Incorporated Paramus, NJ}
}

@article{kartynnik2019,
  title={Real-time Facial Surface Geometry from Monocular Video on Mobile GPUs},
  author={Yury Kartynnik and Artsiom Ablavatski and Ivan Grishchenko and Matthias Grundmann},
  journal={ArXiv},
  year={2019},
  volume={abs/1907.06724},
  url={https://api.semanticscholar.org/CorpusID:196831662}
}

@ARTICLE{Zhang_2016,
  author={Zhang, Kaipeng and Zhang, Zhanpeng and Li, Zhifeng and Qiao, Yu},
  journal={IEEE Signal Processing Letters}, 
  title={Joint Face Detection and Alignment Using Multitask Cascaded Convolutional Networks}, 
  year={2016},
  volume={23},
  number={10},
  pages={1499-1503},
  doi={10.1109/LSP.2016.2603342}}

@article{arriaga2017,
  title={Real-time Convolutional Neural Networks for emotion and gender classification},
  author={Octavio Arriaga and Matias Valdenegro-Toro and Paul-Gerhard Pl{\"o}ger},
  journal={ArXiv},
  year={2017},
  volume={abs/1710.07557},
  url={https://api.semanticscholar.org/CorpusID:6180919}
}

@article{WANG_2026,
title = {Toward noise-resilient retrieval of land surface temperature and emissivity using airborne thermal infrared hyperspectral imagery},
journal = {ISPRS Journal of Photogrammetry and Remote Sensing},
volume = {231},
pages = {532-551},
year = {2026},
issn = {0924-2716},
doi = {https://doi.org/10.1016/j.isprsjprs.2025.10.039},
url = {https://www.sciencedirect.com/science/article/pii/S0924271625004277},
author = {Du Wang and Li-Qin Cao and Yu-Hao Du and Hai-Yang Xiong and Fa-Wang Ye and Yan-Fei Zhong},
}

@article{WANG_2024,
title = {Airborne thermal infrared hyperspectral image temperature and emissivity retrieval based on inter-channel correlated automatic atmospheric compensation and TES},
journal = {Remote Sensing of Environment},
volume = {315},
pages = {114410},
year = {2024},
issn = {0034-4257},
doi = {https://doi.org/10.1016/j.rse.2024.114410},
url = {https://www.sciencedirect.com/science/article/pii/S003442572400436X},
author = {Du Wang and Li-Qin Cao and Lyu-Zhou Gao and Yan-Fei Zhong},
}
\end{document}